\documentclass{article} % For LaTeX2e
\usepackage{nips13submit_e,times}
\usepackage[utf8]{inputenc}
\usepackage[english]{babel}
 
\usepackage{natbib}
\usepackage{hyperref}
\usepackage{url}
\usepackage{graphicx}
\usepackage[]{amssymb}
\usepackage[]{amsmath}

\title{Conditioning of three-dimensional \\ generative adversarial networks \\ for pore and reservoir-scale models}

\author{
Lukas J. Mosser \\
Department of Earth Science and Engineering\\
Imperial College London\\
\texttt{lukas.mosser15@imperial.ac.uk} \\
\And
Olivier Dubrule \\
Department of Earth Science and Engineering\\
Imperial College London\\
\texttt{o.dubrule@imperial.ac.uk} \\
\And
Martin J. Blunt \\
Department of Earth Science and Engineering\\
Imperial College London\\
\texttt{m.blunt@imperial.ac.uk} \\
}

\nipsfinalcopy % Uncomment for camera-ready version

\begin{document}

\maketitle

\begin{abstract}
Geostatistical modeling of petrophysical properties is a key step in modern integrated oil and gas reservoir studies. Recently, generative adversarial networks (GAN) have been shown to be a successful method for generating unconditional simulations of pore- and reservoir-scale models. This contribution leverages the differentiable nature of neural networks to extend GANs to the conditional simulation of three-dimensional pore- and reservoir-scale models. Based on the previous work of \cite{2016arXiv160707539Y}, we use a content loss to constrain to the conditioning data and a perceptual loss obtained from the evaluation of the GAN discriminator network. The technique is tested on the generation of three-dimensional micro-CT images of a Ketton limestone constrained by two-dimensional cross-sections, and on the simulation of the Maules Creek alluvial aquifer constrained by one-dimensional sections.  Our results show that GANs represent a powerful method for sampling conditioned pore and reservoir samples for stochastic reservoir evaluation workflows.\footnote{We have released our code for conditioning at \url{http://github.com/LukasMosser/geogan}}
\end{abstract}

\section{Introduction}

The spatial distribution of rock properties within a reservoir can have a critical impact on hydrocarbon recovery. In recent years a number of geostatistical methods have been developed to generate earth models given sparse information. 

Different approaches exist to model the distribution geological facies and petrophysical properties. Variograms quantify geological and petrophysical variations using so-called two-point statistics \citep{matheron1975random, pyrcz2014geostatistical}. Anisotropic behavior is incorporated by introducing orientation-dependent variograms. Truncated Gaussian simulation obtains facies models by truncation of variogram-based models of Gaussian fields \citep{armstrong2011plurigaussian}. In contrast, multiple-point statistical (MPS) methods \citep{strebelle2002conditional} evaluate the dependency of the facies occurrence at a given location based on statistics available on multi-point templates. These statistics are provided by training images, that represent the conceptual geological knowledge and act as a discrete prior on the geological understanding of the subsurface reservoir. 

Recent methods such as direct sampling \citep{mariethoz2010direct} have led to significant reduction in the computational overhead of MPS methods, allowing rapid sampling of three-dimensional reservoir models. Object-based methods populate model domains with predefined parameterized geometric representations of geobodies. This allows realistic representation of geological features, but the conditioning to well data is challenging when the size of geological objects is large compared to the well spacing.

At the reservoir scale (meters to 10s of kilometers) we have no knowledge of the true subsurface distribution of reservoir properties, except at discrete well locations; on the other hand, at the scale of individual pores of the reservoir rock, direct imaging methods such as micro-computed tomography allow images of the pore-grain structure to be made \citep{blunt2013pore, berg2017industrial}. These images are often limited in size. Where large spatial domains are required for e.g. upscaling tasks, statistical models enable statistical and physical representations of the pore-grain structure. Due to the abundance of two-dimensional thin sections compared to three-dimensional CT measurements, models are often conditioned to match existing two-dimensional images.

Generative adversarial networks (GANs) represent a recent parametric approach developed by \citet{goodfellow2014} to generate realistic samples, given a set of training images. Recently \citet{Mosser17} have shown that GANs are able to generate very realistic stochastic representations of pore-scale structures. \citet{chan2017parametrization} have shown that GANs are able to produce parametric geological representations. \citet{2017arXiv170804975L} incorporate GANs in a Markov Chain Monte-Carlo approach to create representations conditional to dynamic hydraulic data. \citet{LALOY2017387} use conditional MPS simulations as a training set for variational autoencoders to sample conditioned geological representations. \citet{2016arXiv160707539Y} propose a combined "content + perceptual" loss approach and leverage the differentiable and parametric nature of the deep neural networks used to condition GAN simulations to pre-existing data.

\section{Theory}
%\citet{gatys2016image}
Generative adversarial networks (GANs) \citep{goodfellow2014, goodfellowtutorial} are a recent methodology developed in deep learning that allows modeling and sampling from a data distribution represented by a set of training examples. GANs consist of two differentiable functions; a generator $G(\mathbf{z})$ that maps samples obtained from a multivariate standardized normal distribution to an image $\mathbf{x}$ and a discriminator $D(\mathbf{x})$ that takes on the role of a classifier to distinguish between simulations $\mathbf{x}\sim G(\mathbf{z})$ created by the generator and the training images. Both networks are trained in an alternating two-step procedure that optimizes a min-max objective function:
\begin{equation}
 \underset{G}{min} \underset{D}{max}\{\mathbb{E}_{\mathbf{x} \thicksim p_{data}(\mathbf{x})}[log(D(\mathbf{x}))]+\mathbb{E}_{\mathbf{z}\thicksim p_{\mathbf{z}}(\mathbf{z})}[log(1-D(G(\mathbf{z})))]\}\label{objective}
\end{equation}

In this setting, the two functions have distinct and opposing objectives: the discriminator's goal is to distinguish between real training images and samples obtained from the generator, whereas the generator tries to create samples that the discriminator falsely classifies as being a sample of the set of training images. Due to this two-player game between the generator and discriminator the training of GANs is inherently unstable. Using the Wasserstein distance as a surrogate objective function \citep{2017arXiv170107875A} has proven to be a successful way of stabilizing the GAN training process. This contribution uses Wasserstein-GANs combined with a single-sided gradient penalty to train the generator-discriminator pairing \citep{2017arXiv170908894P, gulrajani2017improved}.

Generative adversarial networks were trained on the three-dimensional Maules Creek reservoir dataset \citep{mariethoz2014multiple} and on a gray-scale micro-CT image of a Ketton limestone. The Ketton dataset serves as a pore-scale example for model conditioning \citep{menke2017dynamic}. An overview of the image processing performed on the Ketton dataset can be found in \citet{2017arXiv171202854M}.

The generator and discriminator of each GAN are represented by a deep convolutional neural network (DCGAN) \citep{Radford2016, 2016arXiv161108207J}. Due to the differentiable nature of the deep neural network we can perform gradient-based optimization of the latent vector with respect to an objective function on the output of the GAN generator $G(\mathbf{z})$. 

To constrain models to given conditioning data, the so-called content loss, we use a masked mean-squared error between the generator output $G(\mathbf{z})$ and the conditioning data $\mathbf{y}$, where the mask $\mathbf{M}$ limits the computation of the error to the location of the conditioning data only. 
\begin{equation}
L_{content}=\lVert G(\mathbf{z}) \odot \mathbf{M} - \mathbf{y} \odot \mathbf{M} \rVert
\end{equation}
For binary indicator models, the objective function is the masked cross-entropy between the binary indicator at the conditioning location and the GANs output. Gray-level maps are thresholded at the 0.5 level leading to a binary indicator model sampled from the GAN. 

\citet{2016arXiv160707539Y} showed that only minimizing the content loss does not lead to visual realistic results and therefore introduced a so-called perceptual loss that is given by the discriminator's evaluation of the generator's output $D(G(\mathbf{z}))$. 
\begin{equation}L_{perceptual}=log(1-D(G(\mathbf{z})))
\end{equation}
This perceptual loss evaluates the similarity of patterns observed on the generated samples and the training set. The perceptual loss is therefore added to the content loss in our optimization procedure and weighted by a user defined factor $\lambda$.
\begin{equation}L_{total} = L_{content} + \lambda L_{perceptual}
\end{equation}
For continuous gray-scale images we stop optimization when the content loss is less than 1e-3, whereas for binary indicator models we use a unit accuracy i.e. perfect matching of the indicator variables at the conditioning locations after thresholding as the convergence criteria. 

\begin{equation}Accuracy = \frac{True \ Positives + True \ Negatives}{Positives + Negatives}
\end{equation}

Optimization of the latent vector with regards to the loss is performed using stochastic gradient descent on the latent variable $\mathbf{z}$ until the convergence criteria are met.
%\clearpage
\section{Results}
We evaluate the ability of GANs to generate conditional samples by conditioning the trained GAN networks to lower-dimensional data. Conditioning the three-dimensional output of the Ketton generator network to two-dimensional micro-CT data is performed using orthogonal intersections centered at the origin \citep{Okabe2004}. The location of the orthogonal conditioning planes can be seen (black) below. Although the conditioning data is lower-dimensional, this has a spatial influence on the resulting realizations in the third dimension (Figure 1).

\begin{figure}[!htb]
\centering
\label{fig_1}
\includegraphics[width=\textwidth]{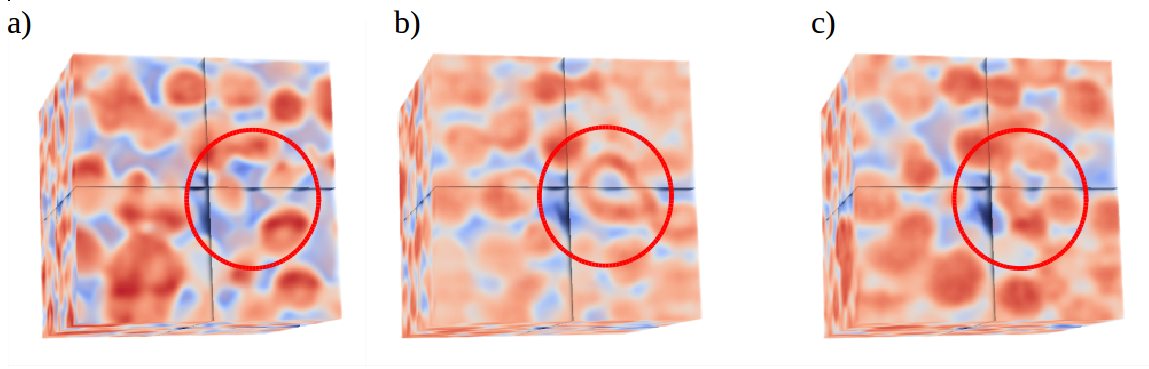}
\caption{A subset of the Ketton limestone training image (Figure 1a) has been used to extract orthogonal two-dimensional cross-sections used as conditioning data (black planes). Two conditioned simulations of a GAN trained on the Ketton dataset are shown (Figure 1b-c). The same lower dimensional conditioning data has a different volumetric expression away from the conditioning planes.}
\end{figure}

For the Maules Creek dataset we condition to a single well in the center of the domains shown in Figure 2.  Conditioning of 1024 Maules Creek three-dimensional simulations was performed on a single GPU in 8 hours. We present the mean and standard deviation of the ensemble of realizations in Figure 2b-c. An elliptical  influence of the conditioning data is observed. Good variation in the samples is shown by the high variance of the conditioned model ensembles. Each conditioned sample matches the indicator data at the well exactly.

\begin{figure}[!htb]
   \centering
   \includegraphics[width=\textwidth]{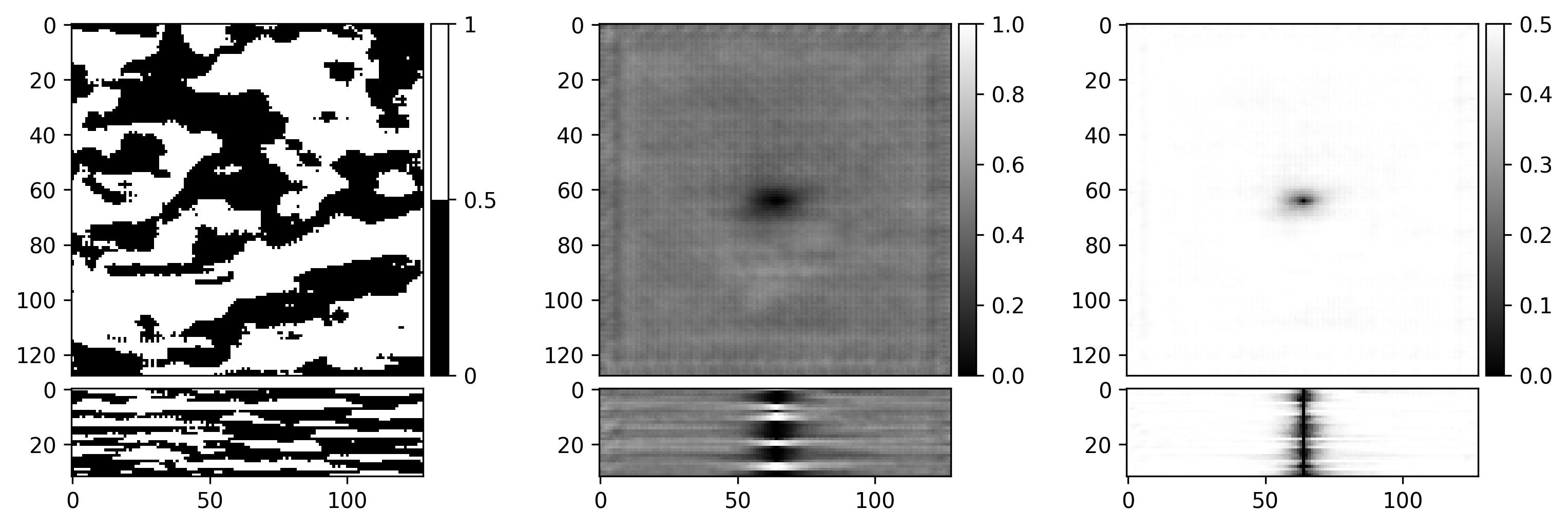}
   \caption{A GAN was used to create simulations of the Maules Creek training image conditional to a single centered well (Figure 2a). Mean and standard deviation maps and cross-sections (Figure 2b-c)  of 1024 conditioned simulations  were created by optimizing the latent vectors of GAN samples trained on the Maules Creek training image. An elliptical region of influence can be observed around the conditional well data. Boundary artifacts observed in the mean of all samples occur due to transposed convolutional layers in the generator architecture.}
 \end{figure}

\section{Conclusions}

Generative adversarial networks are a new powerful machine learning approach for generating three-dimensional simulations of porous media at the reservoir and pore scale. One pore-scale and one reservoir-scale case study have shown that conditional simulation can also be performed, constrained by lower-dimensional well data or cross-sections at the pore scale.

\section{Acknowledgements}

We thank H. Menke for providing the Ketton micro-CT image dataset as well as G. Mariethoz and J. Caers for sharing the Maules Creek training image. O. Dubrule thanks Total S.A. for seconding him as visiting professor at Imperial College London

\bibliography{bibliography}

\begin{thebibliography}{23}
\providecommand{\natexlab}[1]{#1}

\bibitem[{Arjovsky} et~al., 2017]{2017arXiv170107875A}
{Arjovsky}, M., {Chintala}, S. and {Bottou}, L. [2017] {Wasserstein GAN}.
\newblock \emph{arXiv preprint arXiv:1701.07875}.

\bibitem[Armstrong et~al., 2011]{armstrong2011plurigaussian}
Armstrong, M., Galli, A., Beucher, H., Loc'h, G., Renard, D., Doligez, B.,
  Eschard, R. and Geffroy, F. [2011] \emph{Plurigaussian simulations in
  geosciences}.
\newblock Springer Science \& Business Media.

\bibitem[Berg et~al., 2017]{berg2017industrial}
Berg, C.F., Lopez, O. and Berland, H. [2017] Industrial applications of digital
  rock technology.
\newblock \emph{Journal of Petroleum Science and Engineering}, \textbf{157},
  131--147.

\bibitem[Blunt et~al., 2013]{blunt2013pore}
Blunt, M.J., Bijeljic, B., Dong, H., Gharbi, O., Iglauer, S., Mostaghimi, P.,
  Paluszny, A. and Pentland, C. [2013] Pore-scale imaging and modelling.
\newblock \emph{Advances in Water Resources}, \textbf{51}, 197--216.

\bibitem[Chan and Elsheikh, 2017]{chan2017parametrization}
Chan, S. and Elsheikh, A.H. [2017] Parametrization and Generation of Geological
  Models with Generative Adversarial Networks.
\newblock \emph{arXiv preprint arXiv:1708.01810}.

\bibitem[{Goodfellow}, 2017]{goodfellowtutorial}
{Goodfellow}, I. [2017] {NIPS 2016 Tutorial: Generative Adversarial Networks}.
\newblock \emph{arXiv preprint arXiv:1701.00160}.

\bibitem[Goodfellow et~al., 2014]{goodfellow2014}
Goodfellow, I., Pouget-Abadie, J., Mirza, M., Xu, B., Warde-Farley, D., Ozair,
  S., Courville, A. and Bengio, Y. [2014] Generative adversarial nets.
\newblock In: \emph{Advances in Neural Information Processing Systems}.
  2672--2680.

\bibitem[Gulrajani et~al., 2017]{gulrajani2017improved}
Gulrajani, I., Ahmed, F., Arjovsky, M., Dumoulin, V. and Courville, A. [2017]
  Improved training of wasserstein gans.
\newblock \emph{arXiv preprint arXiv:1704.00028}.

\bibitem[{Jetchev} et~al., 2016]{2016arXiv161108207J}
{Jetchev}, N., {Bergmann}, U. and {Vollgraf}, R. [2016] {Texture Synthesis with
  Spatial Generative Adversarial Networks}.
\newblock \emph{arXiv preprint arXiv:1611.08207}.

\bibitem[{Laloy} et~al., 2017]{2017arXiv170804975L}
{Laloy}, E., {H{\'e}rault}, R., {Jacques}, D. and {Linde}, N. [2017] {Efficient
  training-image based geostatistical simulation and inversion using a spatial
  generative adversarial neural network}.
\newblock \emph{ArXiv e-prints}.

\bibitem[Laloy et~al., 2017]{LALOY2017387}
Laloy, E., {H{\'e}rault}, R., Lee, J., Jacques, D. and Linde, N. [2017]
  Inversion using a new low-dimensional representation of complex binary
  geological media based on a deep neural network.
\newblock \emph{Advances in Water Resources}, \textbf{110}, 387 -- 405.

\bibitem[Mariethoz and Caers, 2014]{mariethoz2014multiple}
Mariethoz, G. and Caers, J. [2014] \emph{Multiple-point geostatistics:
  stochastic modeling with training images}.
\newblock John Wiley \& Sons.

\bibitem[Mariethoz et~al., 2010]{mariethoz2010direct}
Mariethoz, G., Renard, P. and Straubhaar, J. [2010] The Direct Sampling method
  to perform multiple-point geostatistical simulations.
\newblock \emph{Water Resources Research}, \textbf{46}(11), 11.

\bibitem[Matheron, 1975]{matheron1975random}
Matheron, G. [1975] \emph{Random sets and integral geometry}.
\newblock Wiley series in probability and mathematical statistics: Probability
  and mathematical statistics. Wiley.

\bibitem[Menke et~al., 2017]{menke2017dynamic}
Menke, H., Bijeljic, B. and Blunt, M. [2017] Dynamic reservoir-condition
  microtomography of reactive transport in complex carbonates: Effect of
  initial pore structure and initial brine pH.
\newblock \emph{Geochimica et Cosmochimica Acta}, \textbf{204}, 267--285.

\bibitem[Mosser et~al., 2017]{Mosser17}
Mosser, L., Dubrule, O. and Blunt, M.J. [2017] Reconstruction of
  three-dimensional porous media using generative adversarial neural networks.
\newblock \emph{Phys. Rev. E}, \textbf{96}, 043309.

\bibitem[{Mosser} et~al., 2017]{2017arXiv171202854M}
{Mosser}, L., {Dubrule}, O. and {Blunt}, M.J. [2017] {Stochastic reconstruction
  of an oolitic limestone by generative adversarial networks}.
\newblock \emph{arXiv preprint arXiv:1712.02854}.

\bibitem[Okabe and Blunt, 2004]{Okabe2004}
Okabe, H. and Blunt, M.J. [2004] Prediction of permeability for porous media
  reconstructed using multiple-point statistics.
\newblock \emph{Phys. Rev. E}, \textbf{70}, 066135.

\bibitem[{Petzka} et~al., 2017]{2017arXiv170908894P}
{Petzka}, H., {Fischer}, A. and {Lukovnicov}, D. [2017] {On the regularization
  of Wasserstein GANs}.
\newblock \emph{arXiv preprint arXiv:1709.08894}.

\bibitem[Pyrcz and Deutsch, 2014]{pyrcz2014geostatistical}
Pyrcz, M.J. and Deutsch, C.V. [2014] \emph{Geostatistical reservoir modeling}.
\newblock Oxford university press.

\bibitem[Radford et~al., 2015]{Radford2016}
Radford, A., Metz, L. and Chintala, S. [2015] Unsupervised representation
  learning with deep convolutional generative adversarial networks.
\newblock \emph{arXiv preprint arXiv:1511.06434}.

\bibitem[Strebelle, 2002]{strebelle2002conditional}
Strebelle, S. [2002] Conditional simulation of complex geological structures
  using multiple-point statistics.
\newblock \emph{Mathematical Geology}, \textbf{34}(1), 1--21.

\bibitem[{Yeh} et~al., 2016]{2016arXiv160707539Y}
{Yeh}, R.A., {Chen}, C., {Yian Lim}, T., {Schwing}, A.G., {Hasegawa-Johnson},
  M. and {Do}, M.N. [2016] {Semantic Image Inpainting with Deep Generative
  Models}.
\newblock \emph{arXiv preprint arXiv:1607.07539}.

\end{thebibliography}

\end{document}